\documentclass[journal]{IEEEtran}
\usepackage{amsmath,amsfonts}
\usepackage{algorithmic}
\usepackage{algorithm}
\usepackage{array}
\usepackage[caption=false,font=normalsize,labelfont=sf,textfont=sf]{subfig}
\usepackage{textcomp}
\usepackage{stfloats}
\usepackage{url}
\usepackage{verbatim}
\usepackage{graphicx}
\usepackage{cite}
\usepackage{times}
\usepackage{epsfig}
\usepackage{amssymb}
\usepackage{fontawesome}
\usepackage{multirow}
\usepackage{tabularx}
\usepackage{color}
\usepackage{mathabx}
\usepackage{pifont}
\begin{document}

\title{Emphasizing Crucial Features for Efficient Image Restoration}

\author{Hu Gao, Bowen Ma, Ying Zhang, Jingfan Yang, Jing Yang and Depeng Dang$^{\dag}$
\thanks{Hu Gao, Depeng Dang
are with the School of
Artificial Intelligence, Beijing Normal University,
Beijing 100000, China (e-mail: gao\_h@mail.bnu.edu.cn, ddepeng@bnu.edu.cn).}}



\maketitle

\begin{abstract}
Image restoration is a challenging ill-posed problem which estimates latent sharp image from  its degraded counterpart. Although the existing methods have achieved  promising performance by designing novelty architecture of module, they ignore the fact that different regions in a corrupted image undergo varying degrees of degradation. In this paper, we propose an efficient and effective framework to adapt to varying degrees of degradation across different regions for image restoration. Specifically, we design a spatial and frequency attention mechanism (SFAM) to emphasize crucial features for restoration. SFAM  consists of two modules: the spatial domain attention module (SDAM) and the frequency domain attention module (FDAM). The SFAM discerns the degradation location through spatial selective attention and channel selective attention in the spatial domain, while the FDAM enhances high-frequency signals to amplify the disparities between sharp and degraded image pairs in the spectral domain. Additionally, to capture global range information, we introduce a multi-scale block (MSBlock) that consists of three scale branches, each containing multiple simplified channel attention blocks (SCABlocks) and a multi-scale feed-forward block (MSFBlock). Finally, we propose our ECFNet, which integrates the aforementioned components into a U-shaped backbone for recovering high-quality images. Extensive experimental results demonstrate the effectiveness of ECFNet, outperforming state-of-the-art (SOTA) methods on both synthetic and real-world datasets. The code and the pre-trained models will be released at~\url{https://github.com/Tombs98/ECFNet}.
\end{abstract}


\section{Introduction}
\IEEEPARstart{I}{mage} degradation commonly arises during image acquisition due to various factors like camera limitations, environmental conditions, and human factors. Image restoration seeks to eliminate this  undesirable degradation to attain high-quality images.
It is a classic example of an ill-posed problem due to the existence of numerous potential solutions. To narrow down the solution space, traditional methods~\cite{karaali2017edge, 2011Image} incorporate prior knowledge and make various assumptions. However, these methods frequently exhibit poor generalization, rendering them impractical for real-world applications.

\begin{figure}[htb] 
	\centering
	\includegraphics[width=1\linewidth]{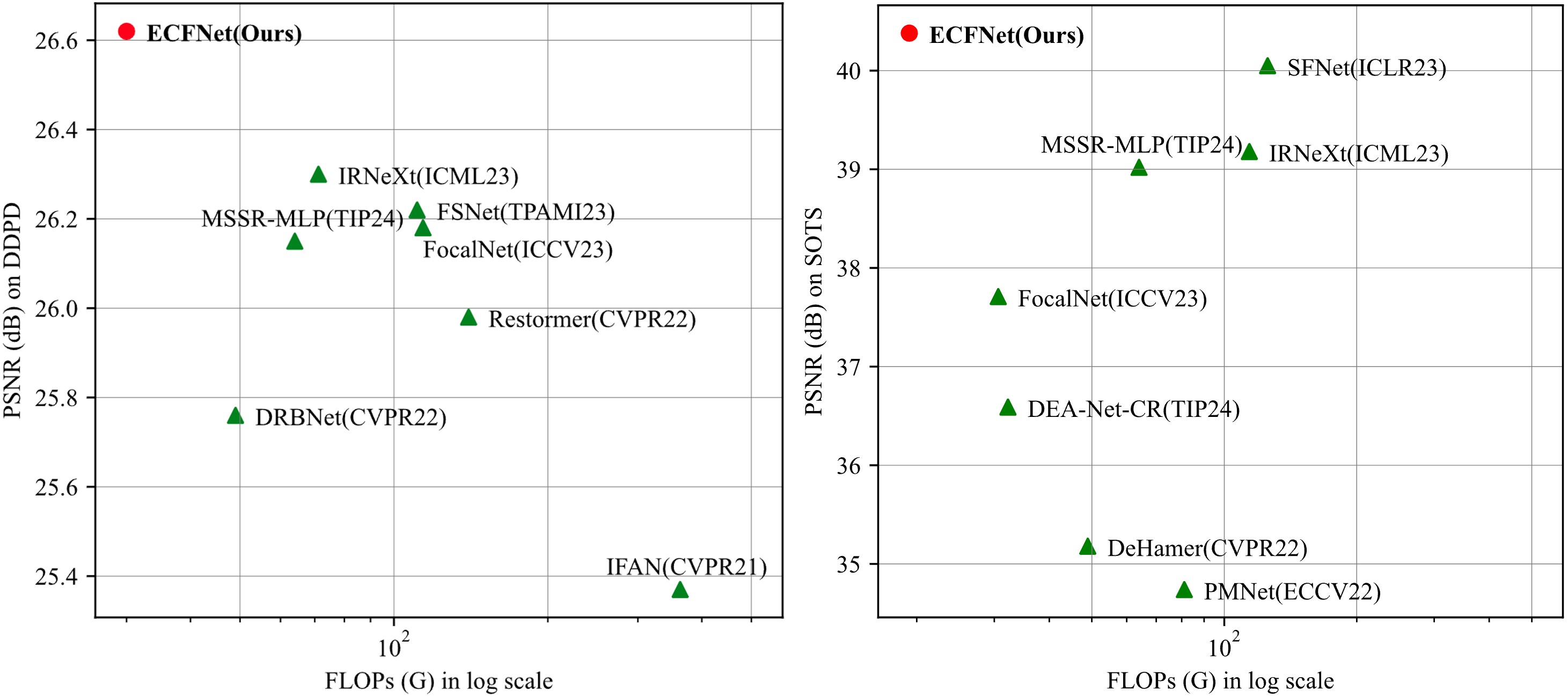}
	\caption{Computational cost vs. PSNR between our ECFNet and other state-of-the-art algorithms. \textbf{Left:} Image defocus deblurring on the DDPD dataset~\cite{DPDNet}, our ECFNet achieve the SOTA performance with up to 57.9\% of cost reduction. \textbf{Right:} Image dehazing on the SOTS dataset~\cite{SOTli2018benchmarking}, our ECFNet achieve the SOTA performance with up to 84.8\% of cost reduction.}
	\label{fig:param}
\end{figure}

With the rapid advancement of deep learning and the availability of large-scale data, convolutional neural networks (CNNs)~\cite{FSNet,deanetchen2024dea,spddkim2022self,focalnetcui2023focal,chen2022simple, 2022Learning, SDLNet} have emerged as the preferred option. They excel in implicitly learning more generalized priors by capturing natural image statistics, thus achieving state-of-the-art (SOTA) performance in image restoration.  
However, the inherent characteristics of CNNs, such as local receptive fields and input content independence, hinder the models' capability to capture long-range dependencies. To address these limitations, several transformer variants~\cite{kong2023efficient,u2former,Zamir2021Restormer,Tsai2022Stripformer, Wang_2022_CVPR} have been employed in image restoration. They have demonstrated superior performance compared to CNN-based methods by leveraging the highly adaptable weights and the global dependency capture ability of self-attention mechanisms. 
Recently, MLP-based methods~\cite{mssrmlphua2024efficient,MAXIMtu2022maxim} have entered the domain of image processing, demonstrating promising results. These methods employ a simple multilayer perceptron instead of self-attention mechanisms and design window-based fully connected layers to reduce computational overhead. 

\begin{figure*}[htb] 
	\centering
	\includegraphics[width=1\linewidth]{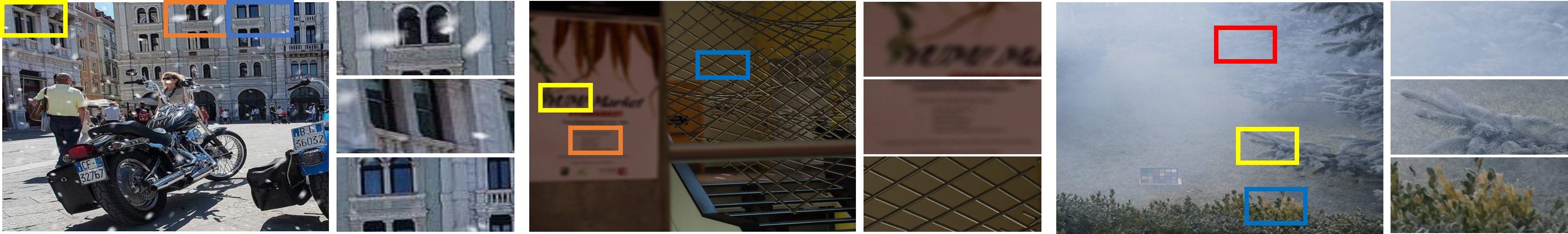}
	\caption{Varying degrees of degradation across different regions. On the left is an image affected by snow from Snow100k~\cite{desnownet}, in the middle is an image affected by blurring from DPDD~\cite{DPDNet}, and on the right is an image affected by haze from O-Haze~\cite{ohazeancuti2018haze}. The yellow box represents the most heavily degrade area, followed by the red box, while the blue box indicates the least degrade region.}
	\label{fig:ques}
\end{figure*}

While the aforementioned methods achieve commendable image restoration performance, they overlook the reality that different regions within a corrupted image often experience degradation to varying extents. As depicted in Figure~\ref{fig:ques}, within the same corrupted image, the area marked by the red box is severely degraded to the point where it's unrecognizable compared to its clear state. Conversely, the blue area exhibits minimal damage or even no degradation at all. When undertakes image restoration, it's evidently unreasonable to treat different regions of a corrupted image as all experiencing the same degree of degradation. 

Considering the analyses presented above, a natural question arises: Is it feasible to devise a network that effectively adapts to varying degrees of degradation across different regions for image restoration? In pursuit of this objective, we propose ECFNet to emphasize critical features for efficient image restoration, incorporating several key components. 1) We design a novel spatial and frequency attention mechanism (SFAM), which consists of two modules: the spatial domain attention module (SDAM) and the frequency domain attention module (FDAM). The SDAM discerns the degradation location through spatial selective attention and channel selective attention in the spatial domain. Additionally, given that frequency disparities between sharp and degraded image pairs primarily reside in the high-frequency components, the FDAM enhances high-frequency signals to amplify the disparities in the spectral domain, aiding in the identification of degraded image regions. 2) To capture global dependencies in images, like symmetry and structural consistency in large objects, we devise multi-scale block (MSBlock). The MSBlock comprises three scale branches, each containing multiple simplified channel attention blocks (SCABlocks) and a multi-scale feed-forward block (MSFBlock). The SCABlock operates on the original resolution features, thereby preserving precise spatial details. Meanwhile, the MSFBlock integrates two multi-scale depth-wise convolution paths to encode multi-scale contextual features. It's important to mention that we exclusively incorporate MSBlocks into the initial  scale of both the encoder and decoder, which focus on the highest resolution features to facilitate multi-scale representation learning. For encoder and decoder at other scales, we introduce multi-scale spatial feature blocks (MSSFBlocks), each consisting of a SCABlock and an MSFBlock. This strategy enables the model to capture global dependency information while also minimizing resource consumption. 3) We integrate the aforementioned modules into a convolutional U-shaped backbone and adopt a coarse-to-fine approach to address challenges in model training, incorporating multiple scales of input and output.
As illustrated in Figure~\ref{fig:param}, our ECFNet model achieves state-of-the-art performance while preserving computational efficiency compared to existing methods.

The main contributions of this work are:
\begin{enumerate}
    \item We propose an efficient and effective framework for image restoration, dubbed ECFNet, which adept at adaptively addressing varying degrees of degradation across different regions within corrupted images.
    
    \item We design a novel spatial and frequency attention mechanism (SFAM), employing the spatial domain attention module (SDAM) to pinpoint degraded regions and the frequency domain attention module (FDAM) to boost high-frequency signals and amplify disparities in the spectral domain, thereby emphasizing crucial features for restoration.
    
    \item We devise multi-scale block (MSBlock), comprising three scale branches, each containing multiple simplified channel attention blocks (SCABlocks) and a multi-scale feedforward block (MSFBlock), to facilitate multi-scale representation learning for image restoration and capture global dependencies.
    
    \item Extensive experiments demonstrate that the proposed ECFNet achieves promising performance compared to state-of-the-art methods across ten synthetic and real-world benchmark datasets encompassing three typical image restoration tasks.
\end{enumerate}

\section{Related Work}
\subsection{Image Restoration}
Images often suffer degradation from equipment and environmental factors during acquisition. Image restoration aims to recover latent sharp images by addressing various degradations such as haze, blur, and snowflakes. Conventional approaches~\cite{2013Unnatural,yang2020single, karaali2017edge, 2011Image, 2011Single} address this ill-posed problem by explicitly integrating priors or hand-crafted features to constrain the solution space to natural images. However, designing such priors is challenging and often lacks generalizability, rendering them impractical for real-world scenarios.
With the rapid advancement of deep learning, numerous works based on deep learning have gained significant popularity in the field of image restoration~\cite{spddkim2022self,deanetchen2024dea,MAXIMtu2022maxim,focalnetcui2023focal,mlptolstikhin2021mlp,chen2022simple,10326458,Zamir2021MPRNet,kong2023efficient,IDT,Zamir2021Restormer,Tsai2022Stripformer,Wang_2022_CVPR,mt10387581,ffanet,IRNeXt}, demonstrating superior performance compared to conventional methods. These approaches are primarily categorized into CNN-based models, Transformer-based models, and MLP-based models.

\subsubsection{CNN-based models}
Instead of manually designing image priors, numerous methods~\cite{deanetchen2024dea,spddkim2022self,IRNeXt,focalnetcui2023focal,chen2022simple,Zamir2021MPRNet,2022Learning,DBGAN,deganv2}  develop a variety of novel network architectures based on CNNs and efficient modules. These approaches implicitly learn more generalized priors by capturing natural image statistics to address the problem of image restoration.
To restore images with precise spatial details and rich contextual information, MPRNet~\cite{Zamir2021MPRNet} divides the image restoration process into multiple stages and conducts feature fusion at various stages using the cross-stage feature fusion mechanism.
FFANet~\cite{ffanet}  devises an innovative attention mechanism to effectively retain information from shallow layers into deep layers for image restoration.
MIRNet-V2~\cite{Zamir2022MIRNetv2} presents a multi-scale architecture that seamlessly integrates contextual information while simultaneously preserving intricate details.
IRNeXt~\cite{IRNeXt} identifies essential properties of image restoration methods and improves multi-scale representation learning.
NAFNet~\cite{chen2022simple} proposes a simplified baseline network by removing or replacing nonlinear activation functions.
DEA-Net~\cite{deanetchen2024dea} introduces  a detail-enhanced attention block to boost the feature learning.
PFONet~\cite{10326458} proposes a lightweight progressive feedback optimization network to effectively handle complex haze scenes.
While these methods outperform hand-crafted prior-based approaches, the inherent characteristics of CNNs, such as local receptive fields and input content independence, limit the models' ability to capture long-range dependencies.

\subsubsection{Transformer-based models}
To capture long-distance dependencies, the transformer architecture~\cite{2017Attention} has recently gained significant traction in image restoration~\cite{kong2023efficient,MRLPFNet,u2former,Zamir2021Restormer,Tsai2022Stripformer, Wang_2022_CVPR,mt10387581,liang2021swinir}.
However, the attention mechanism in Transformers exhibits quadratic computational complexity, rendering it impractical for many image restoration tasks involving high-resolution images. 
To alleviate computational costs, Restormer~\cite{Zamir2021Restormer} and MRLPFNet~\cite{MRLPFNet} apply self-attention across channels instead of spatial dimensions, resulting in linear complexity. Uformer~\cite{Wang_2022_CVPR}, SwinIR~\cite{liang2021swinir}, and U$^2$former~\cite{u2former} compute self-attention using a window-based approach. FFTformer~\cite{kong2023efficient} leverages the properties of the frequency domain to estimate scaled dot-product attention.
Instead of exploring advanced modifications of transformers, this paper tackles the issue of inconsistent degradation in different regions of image restoration by emphasizing crucial features. 

\begin{figure*}[htb] 
	\centering
	\includegraphics[width=1\linewidth]{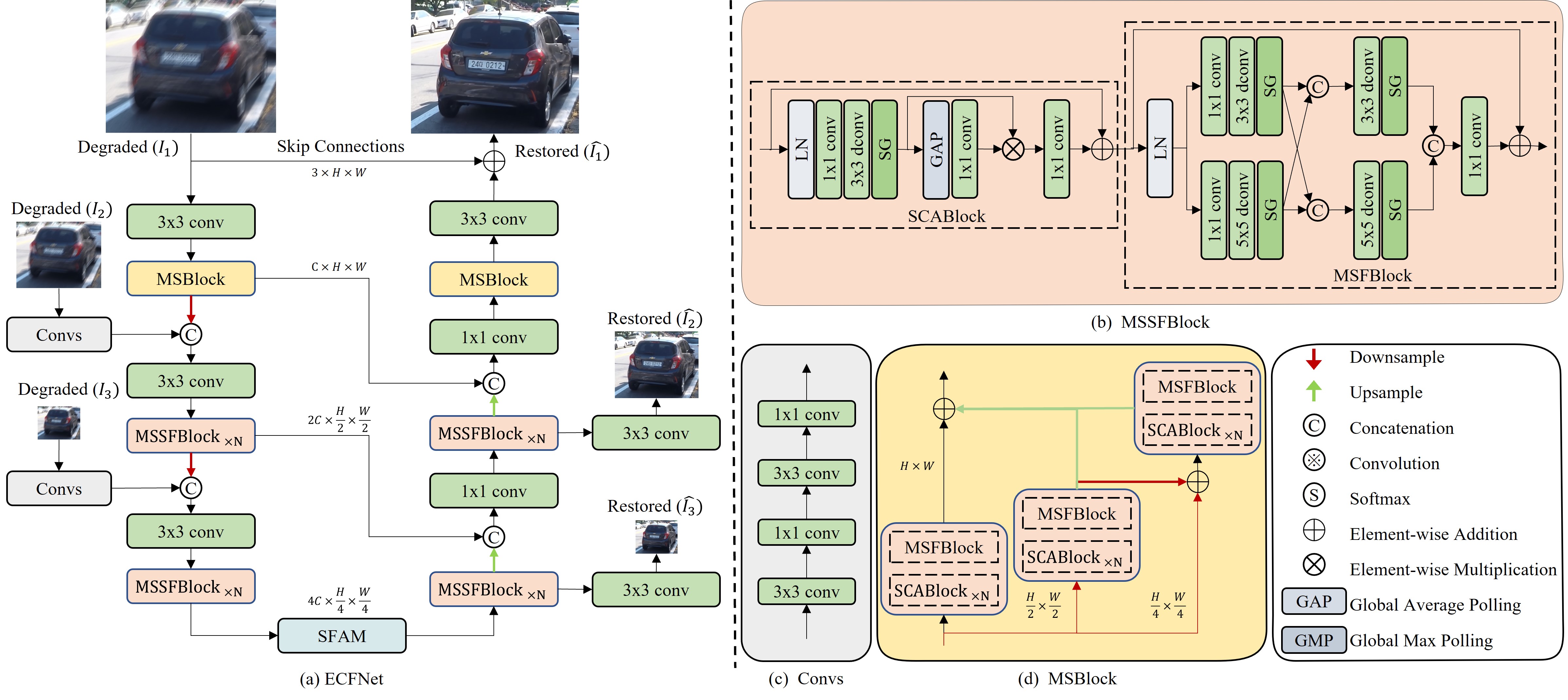}
	\caption{(a) Overall architecture of the proposed ECFNet. (b) Multi-scale spatial feature blocks (MSSFBlock).  (c) ConvS extracts the shallow features for low-resolution degraded images. (d) Multi-scale block (MSBlock), comprising three scale branches.}
	\label{fig:network}
\end{figure*}

\subsubsection{MLP-based models}
Recently, in the field of image processing, there has been a surge in the popularity of simple and efficient MLP-based models that utilize fully connected layers with nonlinear activation functions. One of the key advantages of these models is their ability to reduce the significant computational costs associated with self-attention in transformers. MAXIM~\cite{MAXIMtu2022maxim} introduces a multi-axis gated MLP and a cross-gating block to capture both local and global interactions simultaneously. MSSR-MLP~\cite{mssrmlphua2024efficient} proposes a single-stage multiscale spatial rearrangement multilayer perceptron to gather information at various scales within a local window.

\subsection{Frequency Learning}
Frequency features exhibit significant disparities between damaged images and their corresponding high-quality counterparts. Leveraging this property, frequency learning is employed for image restoration~\cite{fLi2023ICLR,f8803391,fxint2023freqsel,WACAFRN,huang2022WINNet,mf9786841,mf9917526,FSNet, SFNet, 10196308}.  
DeepRFT~\cite{fxint2023freqsel} incorporates Fourier transform to embed kernel-level information into networks designed for image deblurring. 
WACAFRN~\cite{WACAFRN} effectively tackles noise reduction in images by integrating global and local residual blocks. These blocks are guided by both wavelet and adaptive coordinate attention mechanisms, allowing WACAFRN to leverage both global and local features for enhanced performance in noise reduction tasks.
JWSGN~\cite{mf9786841} employs wavelet transform to segregate various frequency components within the image. 
SFNet~\cite{SFNet} and FSNet~\cite{FSNet} introduce a multi-branch dynamic selective frequency module. This module dynamically identifies and selects the most informative components for image restoration, enhancing the overall effectiveness of the restoration process.
AirFormer~\cite{10196308} constructs a frequency-guided Transformer encoder by integrating wavelet-based prior information. This integration guides the extraction of image features, ensuring the preservation of low-frequency components during the encoding process. 
In this paper, we enhance the high-frequency features to amplify the frequency domain disparity between the corrupted image and the high-quality image. This enhancement aids the model in identifying the degraded regions of the image more effectively, leading to improved image restoration outcomes.

\section{Method}
In this section, we first present an overview of the overall pipeline of our ECFNet. Next, we elaborate on the spatial and frequency attention mechanism (SFAM), which includes the spatial domain attention module (SDAM) and the frequency domain attention module (FDAM). We then detail the proposed multi-scale block (MSBlock), which consists of multiple simplified channel attention blocks (SCABlocks) and a multi-scale feed-forward block (MSFBlock). Finally, we describe the loss functions used in our model.

\subsection{Overall Pipeline} 
Our proposed ECFNet, illustrated in Figure~\ref{fig:network}, adheres to the widely embraced encoder-decoder architecture for proficient learning of hierarchical representations. In our methodology, we employ the MSBlock at the first scale, prioritizing high-resolution features. The other two scales predominantly consist of MSSFBlocks. Additionally, we introduce the SFAM into the bottleneck location to accentuate essential features for effective reconstruction. Given a degraded image $\mathbf{I} \in \mathbb R^{H \times W \times 3}$, ECFNet initially applies convolution to obtain shallow features $\mathbf{F_{0}} \in \mathbb R^{H \times W \times C}$ ($H, W, C$ denote the feature map's height, width, and channel number, respectively). These shallow features traverse a three-scale encoder sub-network, progressively reducing resolution while expanding channels. It's worth noting the utilization of multi-input and multi-output mechanisms for enhanced training. The low-resolution degraded images are integrated into the main path through Convs (refer to Figure~\ref{fig:network}(c)) and concatenation, followed by convolution for channel adjustment. The deep features then enter the SFAM to select the most pertinent features, and the resulting deepest features feed into a three-scale decoder, gradually restoring features to their original size. Throughout this process, the encoder features are concatenated with the decoder features to facilitate reconstruction. Finally, we refine features to generate a residual image $\mathbf{R}\in \mathbb R^{H \times W \times 3}$ to which the degraded image is added, yielding the restored image: $\mathbf{\hat{I}} = \mathbf{R} +\mathbf{I}$. It's noteworthy that the two low-resolution results are exclusively used for training purposes. 

\begin{figure*}[htb] 
	\centering
	\includegraphics[width=1\linewidth]{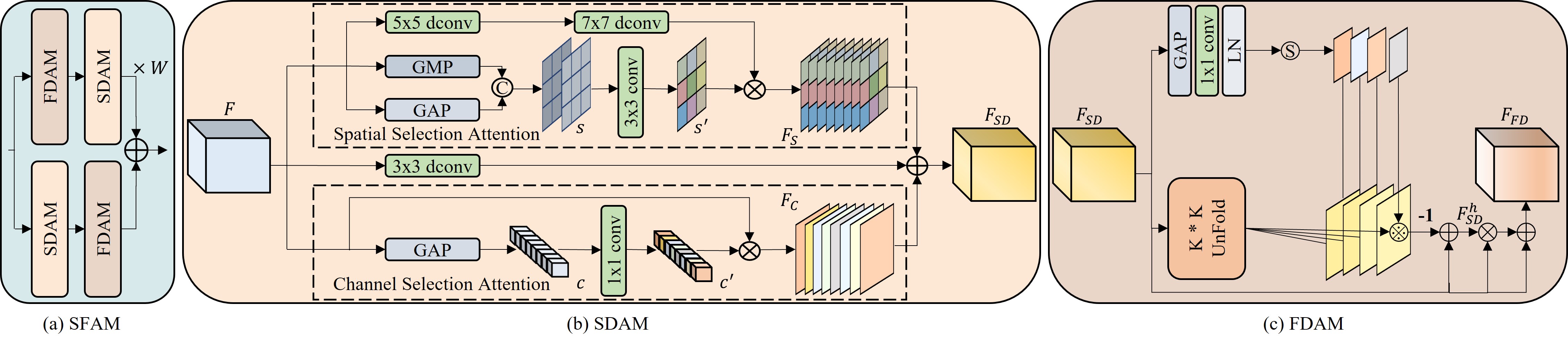}
	\caption{(a) The spatial and frequency attention mechanism (SFAM) that contains: (b) the spatial domain attention module (SDAM) and (c) the frequency domain attention module (FDAM).}
	\label{fig:sfam}
\end{figure*}

\subsection{Spatial and Frequency Attention Mechanism (SFAM)}
Although existing methods~\cite{FSNet,deanetchen2024dea,Zamir2021Restormer,chen2022simple} have achieved impressive results by designing innovative module architectures, they overlook the fact that different regions within a corrupted image experience varying degrees of degradation. As illustrated in Figure~\ref{fig:ques}, within the same corrupted image, the area marked by the red box is severely degraded to the point of being unrecognizable, while the blue area exhibits minimal to no damage. Treating different regions of a corrupted image as if they all suffer the same level of degradation is clearly unreasonable for image restoration tasks. To address this issue, we propose a novel spatial and frequency attention mechanism (SFAM), as shown in Figure~\ref{fig:sfam}, which comprises two modules: the spatial domain attention module (SDAM) and the frequency domain attention module (FDAM). The SFAM identifies degradation locations using spatial selective attention and channel selective attention in the spatial domain. Additionally, considering that frequency differences between sharp and degraded image pairs mainly occur in the high-frequency components, the FDAM enhances high-frequency signals to amplify these differences in the spectral domain, facilitating the identification of degraded regions. The SDAM and FDAM are described below.

\textbf{Spatial Domain Attention Module (SDAM).}
To enable our model to adaptively identify and locate degraded regions in the spatial domain, we design the SDAM. As shown in Figure~\ref{fig:sfam}(b), SDAM primarily consists of spatial selective attention and channel selective attention. For a given feature $\mathbf{F}$, our SDAM enhances the degenerated location features  $\mathbf{F_S}$ in the spatial dimension and the degenerated pattern features $\mathbf{F_C}$ in the channel dimension. These enhanced features are then fused to obtain the important features $\mathbf{F_{SD}}$ in the spatial domain. Formally, the procedures of SDAM can be defined as:
\begin{equation}
\begin{aligned}
	\label{equ:sdam}
 \mathbf{F_{SD}} &= \mathbf{F_{S}} \oplus \mathbf{F_{C}} \oplus f_{3 \times 3}^{dwc}(\mathbf{F})
 \\
 \mathbf{F_{S}} &= SSA(\mathbf{F})
 \\
 \mathbf{F_{C}} &= CSA(\mathbf{F})
\end{aligned}
\end{equation}
where $f_{3 \times 3}^{dwc}$ denotes the $3 \times 3$ depth-wise convolution, $SSA(\cdot)$ and $CSA(\cdot)$ represent the spatial selective attention and channel selective attention, respectively. The spatial selective attention is designed to exploit the focused degradation location features to recalibrate the incoming features $\mathbf{F}$. t first independently applies global average pooling and max pooling operations on $\mathbf{F}$ along the channel dimensions, concatenating the outputs to form a feature map $\mathbf{S} \in \mathbb R^{H \times W \times 2}$. The map $\mathbf{S}$ is passed through a convolution to generate the general feature map $\mathbf{S^{'}} \in \mathbb R^{H \times W \times 1}$, which contains degradation locations to focus on. Finally, $\mathbf{S^{'}}$ is used  to  rescale $\mathbf{F}$ to degenerated location features  $\mathbf{F_S}$ in the spatial dimension. This process is expressed as follows:
\begin{equation}
\begin{aligned}
	\label{equ:sdamsp}
 \mathbf{S^{'}} &= f_{3 \times 3}^{dwc}([GAP(\mathbf{F}), GMP(\mathbf{F})])
 \\
 \mathbf{F_S} &=  \mathbf{S^{'}} \otimes f_{7 \times 7}^{dwc}(f_{5 \times 5}^{dwc}(\mathbf{F}))
\end{aligned}
\end{equation}
where $[\cdot]$ represents the channel-wise concatenation, GAP and GMP denote the global average pooling and the global max pooling, respectively.

Given that each channel exhibits different degradation patterns, we further generate a channel-wise representation using channel selective attention. This process begins by encoding global context through global average pooling across the spatial dimensions to yield a feature descriptor $\mathbf{C} \in \mathbb R^{1 \times 1 \times C}$. This descriptor $\mathbf{C}$ is then passed through a convolutional layer to generate the channel attention score $\mathbf{C^{'}}$. Finally, we obtain the channel-wise degenerated pattern representation $\mathbf{F_C}$ by rescaling $\mathbf{F}$ with the channel attention score $\mathbf{C^{'}}$. Formally, this process can be defined as:
\begin{equation}
	\label{equ:sdamca}
 \mathbf{F_C} = f_{1 \times 1}^{c}(GAP(\mathbf{F})) \otimes \mathbf{F}
\end{equation}
where  $f_{1 \times 1}^c$ denotes the  $1 \times 1$ convolution.
 
\textbf{Frequency Domain Attention Module (FDAM).} 
Considering that frequency differences between sharp and degraded images are mainly present in the high-frequency components, we enhance these high-frequency signals to highlight the disparities in the spectral domain, aiding in the identification of degraded regions. Based on this principle,  we design the FDAM. As illustrated in Figure~\ref{fig:sfam}(c), FDAM  first generates low-frequency maps dynamically using a learnable filter, then derives the complementary high-frequency features by subtracting these low-frequency signals from the input.
Specifically, given an input feature $\mathbf{F_{SD}} \in \mathbb{R}^{H \times W \times C}$, we initially employ GAP, convolution, Layer Normalization (LN), and a softmax activation function  as outlined below:
\begin{equation}
\label{eq:fl}
\mathbf{F^l} = Softmax(LN( f_{1 \times 1}^c(GAP(\mathbf{F_{SD}} ))))
\end{equation}
We reshape $\mathbf{F^l}$ from $\mathbb R^{1 \times 1 \times P}$ to $\mathbb R^{g \times k \times k}$, where $g = \frac{P}{k \times k}$. We then apply the softmax activation function to each $k \times k$ filter  to ensure $\mathbf{F^l}_{g,i,j} \geq 0$ and $\sum_{i,j} \mathbf{F^l}_{g,i,j} = 1$, where $i$ and $j$ denote the indices of the rows and columns in $\mathbf{F^l}_{g}$. Each $g$ of $\mathbf{F^l}_{g}$ thus acts as a low-pass filter.Unlike the bilateral filter, which weighs spatial and color similarity based on the exponential Euclidean distance, our filter is spatially variant and can more adaptively model the main structures.  Next, we use $k \times k$ kernel sizes to unfold the input feature $\mathbf{F_{SD}}$. The resulting output is convolved with the low-pass filter $\mathbf{F^l}$, and then reshaped back to the original feature map size to obtain the final low-frequency feature map $\mathbf{F_{SD}^l}$. This process can be expressed as:
\begin{equation}
	\label{equ:fl}
	\mathbf{F_{SD}^l} =  \mathbf{F^l} \ocoasterisk UnFold(\mathbf{F_{SD}})
\end{equation}

To obtain the high-frequency feature map $\mathbf{F_{SD}^h}$, we subtract the resulting low-frequency feature map $\mathbf{F_{SD}^l}$  from the input feature $\mathbf{F_{SD}}$, which is expressed as:
\begin{equation}
	\label{equ:fh}
	\mathbf{F_{SD}^h}=  \mathbf{F_{SD}} - \mathbf{F_{SD}^l}
\end{equation}

Finally, we utilize the features $\mathbf{F_{SD}^h}$ to emphasize the genuinely important degradation regions $ \mathbf{F_{FD}}$ for reconstruction as follows:
\begin{equation}
\label{equ:sff2}
    \mathbf{F_{FD}} = \mathbf{F_{SD}^h} \otimes \mathbf{F_{SD}} + \mathbf{F_{SD}}
\end{equation}

To eliminate the impact of the order of SDAM and FDAM on the results, SFAM employs a parallel branch structure for both orders and uses trainable channel-wise parameters to reweigh. It is worth mentioning that we propose increasing the weight of the branch where SDAM precedes FDAM . The overall process is as follows:
\begin{equation}
\begin{aligned}
\label{equ:sfor}
    \mathbf{SFAM_1} &= SDAM(FDAM(\mathbf{F}))
    \\
    \mathbf{SFAM_2} &= FDAM(SDAM(\mathbf{F}))
    \\
    \mathbf{SFAM} &= W \times \mathbf{SFAM_1} + \mathbf{SFAM_2}
\end{aligned}
\end{equation}
where $W$ denotes the learnable parameters, which are directly optimized by backpropagation and initialized as \textbf{1}.  

\subsection{Multi-scale Block (MSBlock)}
Natural images contains a global span of features, such as global symmetry, multi-scale pattern repetition, same-scale texture similarity, and structural consistency in large objects and content. These global features are essential for image restoration tasks. Rather than improving the transformer architecture, we explored a multi-scale method based on CNNs and proposed multi-scale blocks (MSBlocks) to capture these global features through multi-scale learning. 

The architecture of MSBlock is shown in Figure~\ref{fig:network}(d).  Given an input tensor $\mathbf{X} \in \mathbb{R}^{H \times W}$, with the channel dimension omitted for simplicity, our MSBlock uses different downsampling ratios to convert $\mathbf{X}$ into distinct scale spaces. These various scale features are then processed through three branches of the MSBlock. Each branch consists of multiple simplified channel attention blocks (SCABlocks) that operate on the original resolution features to preserve precise spatial details, and a multi-scale feed-forward block (MSFBlock) that integrates two multi-scale depth-wise convolution paths to encode multi-scale contextual features. For the $i_{th} (i \in {1, 2, 3})$ branch, the output features can be obtained by:
\begin{equation}
\begin{aligned}
\label{eq:0msbb1}
    \mathbf{X_1} &= MSFB(SCAB_1(...SCAB_N(\mathbf{X})...))
    \\
    \mathbf{X_2} &= MSFB(SCAB_1(...SCAB_N(\mathbf{X}\downarrow_{2}...))
    \\
    \mathbf{X_3} &= MSFB(SCAB_1(...SCAB_N(\mathbf{X}\downarrow_{4} \oplus 
 \mathbf{X_2}\downarrow_{2})...))
\end{aligned}
\end{equation}
where $\downarrow_{j}$ denotes downsampling with a ratio of $j$. Finally, the output features from all branches are unified to the same feature size through dynamic upsampling and then fused as follows:
\begin{equation}
\label{eq:0msbfu}
    \mathbf{\hat{X}} = \mathbf{X_1} \oplus \mathbf{X_2}\uparrow_{2} \oplus \mathbf{X_3}\uparrow_{4}
\end{equation}
where $\uparrow_{j}$ denotes dynamic upsampling with a ratio of $j$.

\textbf{Simplified Channel Attention Block (SCABlock).}
Inspired by NAFNet~\cite{chen2022simple}, we designed a simplified channel attention block(SCABlock) to reduce computational resource consumption. This module uses a simple gating mechanism in place of a nonlinear activation function, while still effectively integrating global and channel information. As shown in Figure~\ref{fig:network}(b), given an input tensor $\mathbf{X} \in \mathbb{R}^{H \times W \times C}$, the precise spatial detail features $\mathbf{X_s}$ can be obtained by:
\begin{equation}
\begin{aligned}
	\label{equ:msbsca}
	\mathbf{X^{'}} &= SG(f_{3 \times 3}^{dwc} (f_{1 \times 1}^c(LN(\mathbf{X}))))
 \\
	\mathbf{X_s} &= f_{1 \times 1}^c(\mathbf{X^{'}} \otimes f_{1 \times 1}^c(GAP(\mathbf{X^{'}}))) \oplus \mathbf{X}
\\
    SG(\mathbf{F^{'}}) &= \mathbf{F^{'}_1} \otimes \mathbf{F^{'}_2} 
\end{aligned}
\end{equation}
where  SG denotes the simple gating mechanism, which starts by splitting a feature into two along the channel dimension. Then computes these two features using a linear gate.

\textbf{Multi-scale Feed-forward Block (MSFBlock)}
Most prior work~\cite{chen2022simple,chu2022nafssr} focus on enhancing locality by incorporating single-scale depth-wise convolutions into feed-forward neural networks. However, these methods overlook the multi-scale pattern repetition inherent in images. To address this, we design a multi-scale feed-forward block (MSFBlock) to improve multi-scale representation by utilizing depth-wise convolutional branches at two different scales, thereby enhancing the ability to capture global dependencies. Given the output features $\mathbf{X_s}$ from SCABlock, after layer normalization, we feed them into two parallel branches to obtain context-rich features $\mathbf{X_c}$. The entire procedure can be formulated as follows:
\begin{equation}
\begin{aligned}
	\label{equ:1msbmsf}
 \mathbf{X^{t}}&=  SG(f_{3 \times 3}^{dwc}(f_{1 \times 1}^c (LN(\mathbf{X^{s}}))))
\\
 \mathbf{X^{b}}&=  SG(f_{5 \times 5}^{dwc}(f_{1 \times 1}^c (LN(\mathbf{X^{s}}))))
 \\
 \mathbf{X^{'}_c} & = [SG(f_{3 \times 3}^{dwc}([\mathbf{X^{t}}, \mathbf{X^{b}}])), SG(f_{5 \times 5}^{dwc}([\mathbf{X^{b}}, \mathbf{X^{t}}]))]
 \\
\mathbf{X_c} &=  f_{1 \times 1}^c(\mathbf{X^{'}_c}) \oplus \mathbf{X^{s}}
\end{aligned}
\end{equation}

It's important to note that we integrate MSBlocks exclusively into the initial scale of both the encoder and decoder to handle the highest resolution features, thereby enhancing multi-scale representation learning. For the subsequent scales in the encoder and decoder, we utilize multi-scale spatial feature blocks (MSSFBlocks). This approach allows the model to effectively capture global dependency information while optimizing resource usage. As illustrated in  Figure~\ref{fig:network}(b), the MSSFBlock comprises a SCABlock and an MSFBlock. Given the input features from the $(l-1)$ -th block $\mathbf{F_{l-1}}$, the procedures of MSSFBlock can be defined as:
\begin{equation}
\begin{aligned}
\label{eq:mssfblock}
\mathbf{F_{l-1}^{'}} & = \mathbf{F_{l-1}} \oplus SCAB(\mathbf{F_{l-1}})
\\
\mathbf{F_{l}} &= \mathbf{\mathbf{F_{l-1}^{'}}}  \oplus MSFB(\mathbf{F_{l-1}^{'}})
\end{aligned}
\end{equation}

\subsection{Loss Function}
To facilitate the selection process across both spatial and frequency domains, we optimize the proposed network ECFNet with the following loss function:
\begin{equation}
\begin{aligned}
\label{eq:loss1}
L &= \sum_{i=1}^{4}(L_{c}(\hat{I_i},\overline I_i)  + \delta L_{e}(\hat{I_i},\overline I_i) + \lambda L_{f}(\hat{I_i},\overline I_i))
\\
L_{c} &= \sqrt{||\hat{I_i} -\overline I_i||^2 + \epsilon^2}
\\
L_{e} &= \sqrt{||\triangle \hat{I_i} - \triangle \overline I_i||^2 + \epsilon^2}
\\
L_{f} &= ||\mathcal{F}(\hat{I}_i)-\mathcal{F}(\overline I_i)||_1
\end{aligned}
\end{equation}

where $i$ denotes the index of input/output images at different scales, $\overline I_i$ denotes the target images and $L_{c}$ is the  Charbonnier loss with constant $\epsilon$ empirically set to $0.001$ for all the experiments. $L_{e}$ is the edge loss, where $\triangle$ represents the  Laplacian operator. $L_{f}$  denotes the frequency domains loss, where $\mathcal{F}$ represents fast Fourier transform, and the parameters $\lambda$ and $\delta$ control the relative importance of loss terms, which are set to $0.1$ and $ 0.05$ as in~\cite{Zamir2021MPRNet,FSNet}, respectively.

\section{Experiments}
In this section, we outline the experimental settings and subsequently present both qualitative and quantitative comparisons between ECFNet and other state-of-the-art methods. Following this, we conduct ablation studies to validate the effectiveness of our approach. The best and second-best results in tables are highlighted in \textbf{bold} and \underline{underlined} formats, respectively.

\subsection{Experimental Settings}
\subsubsection{Datasets}
In this section, we will introduce the datasets used and provide details regarding the training configurations.

\textbf{Image Dehazing.} 
We assess the performance of our ECFNet on both daytime and nighttime datasets. The daytime datasets encompass synthetic data from RESIDE~\cite{SOTli2018benchmarking} and real-world data from NH-HAZE~\cite{dhhazeancuti2020nh}, Dense-Haze~\cite{densehazeancuti2019dense}, and O-HAZE~\cite{ohazeancuti2018haze}. The RESIDE dataset contains two training subsets: the indoor training set (ITS) and the outdoor training set (OTS), along with a synthetic objective testing set (SOTS). The ITS consists of 13,990 hazy images generated from 1,399 sharp images, while the OTS comprises 313,950 hazy images derived from 8,970 clean images.
Our model is trained separately on the ITS and OTS datasets, and subsequently tested on their corresponding test sets, namely SOTS-Indoor and SOTS-Outdoor, each comprising 500 paired images. For training on RESIDE-Outdoor, the model undergoes 30 epochs with an initial learning rate set to $1e^{-4}$ and a batch size of 16. On the other hand, for training on RESIDE-Indoor, the model is trained for 1,000 epochs. Furthermore, we incorporate real-world datasets including NH-HAZE~\cite{dhhazeancuti2020nh}, Dense-Haze~\cite{densehazeancuti2019dense}, and O-HAZE~\cite{ohazeancuti2018haze}. NH-HAZE~\cite{dhhazeancuti2020nh} and Dense-Haze~\cite{densehazeancuti2019dense} each comprise 55 paired images, while O-HAZE~\cite{ohazeancuti2018haze} consists of 45 image pairs. For these real-world datasets, models are trained for 5,000 epochs following~\cite{FSNet}, utilizing 600 × 800 patches with an initial learning rate set to $2e^{-4}$ and a batch size of 2. Additionally, we evaluate our models on the nighttime dataset NHR~\cite{nhrzhang2020nighttime}, which encompasses 16,146 and 1,794 image pairs for training and evaluation, respectively. The model is trained for 300 epochs with an initial learning rate of $1e^{-4}$ and a batch size of 8.

\textbf{Single-Image Defocus Deblurring.}
We employ the DPDD dataset~\cite{DPDNet} to demonstrate the efficacy of our method, following the methodology of recent approaches~\cite{FSNet,Zamir2021Restormer}. This dataset comprises images from 500 indoor/outdoor scenes captured using a DSLR camera. Each scene contains four defocused input images along with a corresponding all-in-focus ground-truth image. These images are labeled as right view, left view, center view, and the all-in-focus ground truth. The DPDD dataset is partitioned into training, validation, and testing sets, consisting of 350, 74, and 76 scenes, respectively. ECFNet is trained using the center view images as input, with loss computed between the outputs and corresponding ground-truth images. We adopt a training strategy similar to that of IRNeXt~\cite{IRNeXt}.

\textbf{Image Desnowing.}
For desnowing evaluation, we utilize three datasets: Snow100K~\cite{desnownet}, SRRS~\cite{JSTASRchen2020jstasr}, and CSD~\cite{HDCW-Netchen2021all}. Snow100K~\cite{desnownet} comprises 50,000 image pairs for training and 50,000 for evaluation. SRRS~\cite{JSTASRchen2020jstasr} contains 15,005 image pairs for training and 15,005 for evaluation. CSD~\cite{HDCW-Netchen2021all} consists of 8,000 image pairs for training and 2,000 for evaluation.
To maintain consistency with the training strategy of the previous algorithm~\cite{FSNet}, we randomly sample 2,500 image pairs from the training set for training and 2,000 images from the testing set for evaluation. Models are trained for 2,000 epochs on each dataset.

\subsubsection{Training details}
For different tasks, we train separate models, and unless specified otherwise, the following parameters are employed. The models are trained using the Adam optimizer~\cite{2014Adam} with parameters $\beta_1=0.9$ and $\beta_2=0.999$. The initial learning rate is set to $8 \times 10^{-4}$ and gradually reduced to $1 \times 10^{-6}$ using the cosine annealing strategy~\cite{2016SGDR}. The batch size is chosen as $32$, and patches of size $256 \times 256$ are extracted from training images. Data augmentation involves horizontal and vertical flips.  Depending on the task complexity, we scale the model by varying the number of N in Figure~\ref{fig:network}, i.e., 8 for deblurring/desnowing and 4 for dehazing. In our ECFNet, we only deploy MSBlock in the first scale of encoder/decoder.

\subsection{Experimental Results}

\subsubsection{Image Dehazing}
We assess the performance of our ECFNet on both daytime and nighttime datasets.

\begin{figure*}[htb] 
	\centering
	\includegraphics[width=1\linewidth]{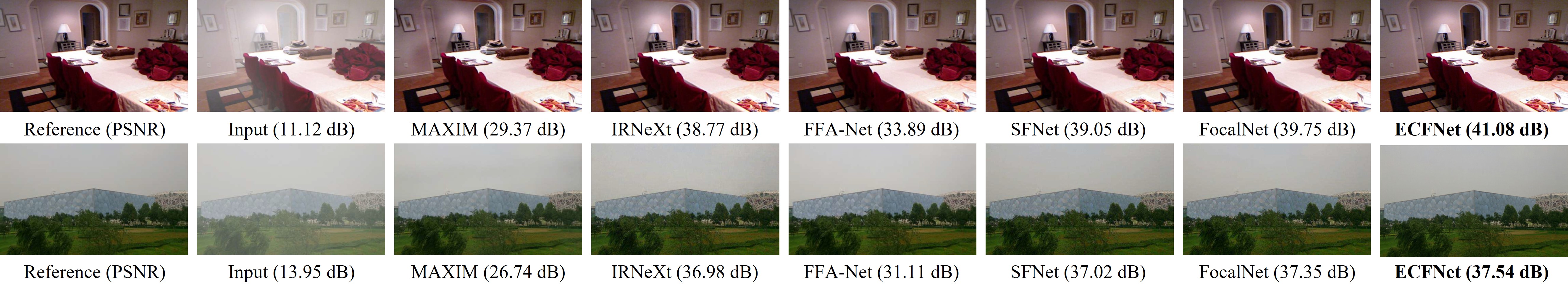}
	\caption{Image dehazing comparisons on the SOTS dataset~\cite{SOTli2018benchmarking}. The top image is obtained from SOTS-Indoor while the bottom one is from SOTS-Outdoor. }
	\label{fig:sot}
\end{figure*}

\begin{figure}[htb] 
	\centering
	\includegraphics[width=1\linewidth]{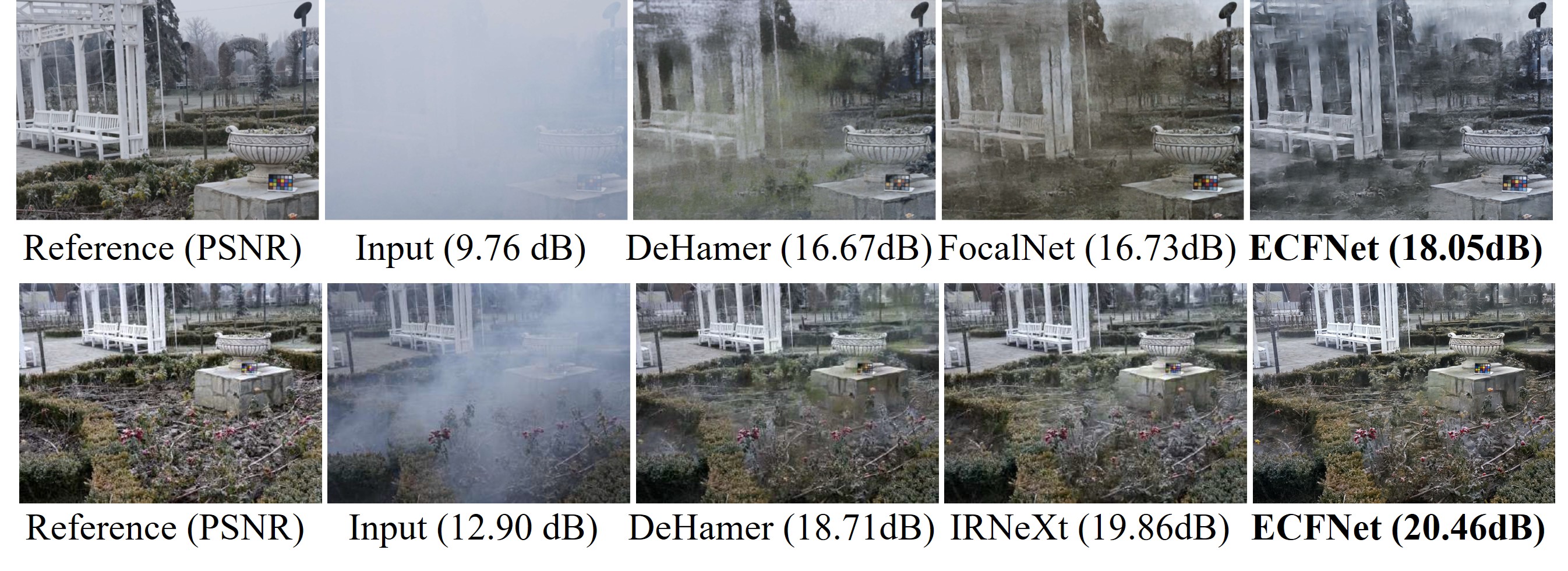}
	\caption{ Image dehazing comparisons on the real world dataset. The top image is obtained from Dense-Haze dataset~\cite{densehazeancuti2019dense} while the bottom one is from NH-Haze dataset~\cite{dhhazeancuti2020nh}.}
	\label{fig:dn}
\end{figure}

\textbf{Daytime datasets.} We present the quantitative performance of various image dehazing methods on both synthetic (RESIDE~\cite{SOTli2018benchmarking}) and real-world datasets (NH-HAZE~\cite{dhhazeancuti2020nh}, Dense-Haze~\cite{densehazeancuti2019dense}, and O-HAZE~\cite{ohazeancuti2018haze}). The quantitative results for the synthetic RESIDE dataset are displayed in Table~\ref{tab:sot}. Overall, our ECFNet demonstrates superior performance in both indoor and outdoor scenes. Specifically, our method surpasses DEA-Net-CR~\cite{deanetchen2024dea} by 0.65 dB PSNR on the SOTS-Indoor dataset~\cite{SOTli2018benchmarking} and by 3.79 dB PSNR on the SOTS-Outdoor dataset~\cite{SOTli2018benchmarking}, while utilizing only 56\% of the parameters and 59\% of the FLOPs. Furthermore, on the SOTS-Indoor dataset~\cite{SOTli2018benchmarking}, our model achieves a notable improvement of 2.53 dB PSNR over the Transformer model DehazeFormer-L~\cite{dehazeformersong2023vision}, with 92\% fewer parameters and 93\% fewer FLOPs. Figure~\ref{fig:sot} shows that our results are visually closer to the ground truth. Additionally, as demonstrated in Table~\ref{tab:realhaze}, our method generalizes well to more challenging real-world scenarios, achieving the best performance. Specifically, on the NH-Haze dataset~\cite{dhhazeancuti2020nh}, ECFNet achieves a significant gain of 0.72 dB PSNR over DeHamer~\cite{Dehamerguo2022image}. The visual results produced by several dehazing methods are illustrated in Figure~\ref{fig:dn}. Our method is more effective in removing haze compared to other frameworks.

\begin{figure}[htb] 
	\centering
	\includegraphics[width=1\linewidth]{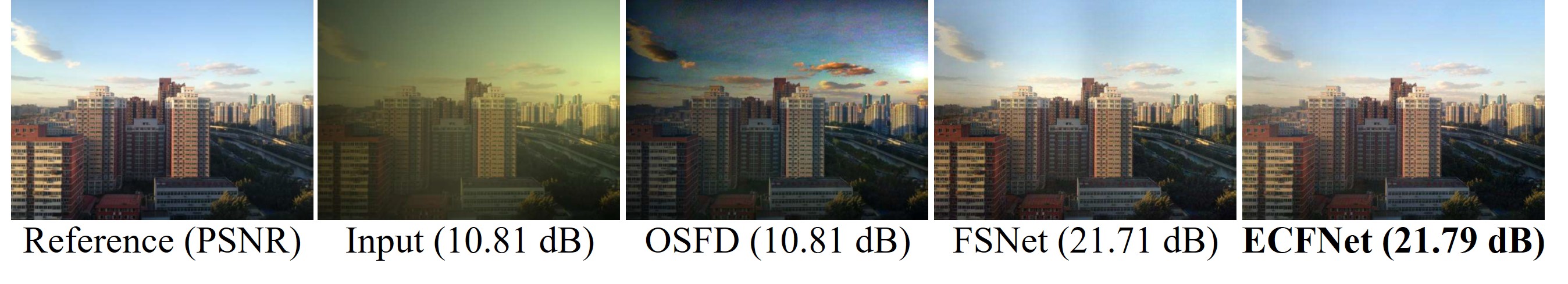}
	\caption{Nighttime dehazing comparisons on the NHR dataset~\cite{nhrzhang2020nighttime}.}
	\label{fig:nhr}
\end{figure}

\textbf{Nighttime datasets.} 
In addition to daytime dehazing datasets, we also evaluate the effectiveness of our ECFNet on the nighttime dehazing dataset NHR~\cite{nhrzhang2020nighttime}. As shown in Table~\ref{tab:nhr}, ECFNet achieves the best performance. Figure~\ref{fig:nhr} demonstrates that our ECFNet is more effective at removing haze blur in nighttime scenes.

\begin{table}
    \centering
        \caption{Quantitative comparisons on the on the synthetic dehazing datasets: SOTS-Indoor and SOTS-Outdoor~\cite{SOTli2018benchmarking}.}
    \label{tab:sot}
    \resizebox{\linewidth}{!}{
    \begin{tabular}{c|cc|cc|cc}
    \hline
    \multicolumn{1}{c|}{} & \multicolumn{2}{c|}{SOTS-Indoor}  & \multicolumn{2}{c|}{SOTS-Outdoor} & Params & FLOPs
    \\
   Methods & PSNR $\uparrow$ & SSIM $\uparrow$  & PSNR $\uparrow$ & SSIM $\uparrow$ & (M) & (G)
   \\
   \hline
   \hline
       GridDehazeNet~\cite{Griddehaliu2019griddehazenet} &32.16 &0.984 &30.86 &0.982 &0.96 &21.49
       \\
MSBDN~\cite{MSBDNdong2020multi} &33.67 &0.985 &33.48 &0.982 &31.35 &41.54
\\
FFA-Net~\cite{FFANetqin2020ffa}&36.39 &0.989 &33.57 &0.984 &4.46 &287.80
\\
AECR-Net~\cite{AECRNetwu2021contrastive} &37.17 &0.990 &- &- &2.61 &52.20
\\
DeHamer~\cite{Dehamerguo2022image} &36.63 &0.988 &35.18 &0.986 &132.45 &48.93
\\
DehazeFormer-L~\cite{dehazeformersong2023vision} &40.05 &\underline{0.996} &- &- &25.44 &279.70
\\
MAXIM~\cite{MAXIMtu2022maxim}&38.11 &0.991 &34.19 &0.985 &14.10 &216.00
\\
PMNet~\cite{PMNetye2022perceiving} &38.41 &0.990 &34.74 &0.985 &18.90 &81.13
\\
SPDD~\cite{spddkim2022self} & 38.50 & 0.992 & 34.50 & 0.987 &- & -
\\
SFNet~\cite{SFNet} & 41.24 &\underline{0.996} &\underline{40.05} &\textbf{0.996} & 13.27 &125.43
\\
FocalNet~\cite{focalnetcui2023focal}& 40.82 &\underline{0.996} &37.71 &\underline{0.995}&3.74 &30.63
\\
IRXNext~\cite{IRNeXt}&41.21 &\underline{0.996} &39.18 &\textbf{0.996} &5.46 & 114.79
\\
DEA-Net-CR~\cite{deanetchen2024dea} & \underline{41.31} &0.995 & 36.59 & 0.990 & 3.65 & 32.23
\\
         \hline
\textbf{ECFNet(Ours)} & \textbf{41.96} & \textbf{0.997} &\textbf{40.38} &0.993 &2.06 & 19.23
         \\
         \hline
    \end{tabular}}
\end{table}

\begin{table}
    \centering
        \caption{Quantitative comparisons on the on the real-world dehazing datasets: Dense-Haze~\cite{densehazeancuti2019dense}, NH-Haze~\cite{dhhazeancuti2020nh} and O-Haze~\cite{ohazeancuti2018haze}.}
    \label{tab:realhaze}
    \resizebox{\linewidth}{!}{
    \begin{tabular}{c|c|c|c}
    \hline
    Methods & Dense-Haze  & NH-Haze & O-Haze
    \\
   \hline
   \hline
MSBDN~\cite{MSBDNdong2020multi}  &15.37  &19.23  &24.36 
\\
FFA-Net~\cite{FFANetqin2020ffa} &14.39 &19.87  &22.12 
\\
AECR-Net~\cite{AECRNetwu2021contrastive}  &15.80  &19.88  &- 
\\
DeHamer~\cite{Dehamerguo2022image}  &16.62  &\underline{20.66}  &- 
\\
PMNet~\cite{PMNetye2022perceiving} & 16.79  &20.42  &24.64 
\\
FSNet~\cite{FSNet} &  17.13  &20.55  & - 
\\
IRXNext~\cite{IRNeXt}&\underline{17.60}  &20.55 &-
\\
FocalNet~\cite{focalnetcui2023focal}& 17.07  &20.43 &\underline{25.50} 
\\  
         \hline
\textbf{ECFNet(Ours)} & \textbf{18.15}  &\textbf{21.38} &\textbf{25.87} 
         \\
         \hline
    \end{tabular}}
\end{table}


\begin{table}
    \centering
     \caption{ Nighttime image dehazing results on NHR dataset~\cite{nhrzhang2020nighttime}.}
    \label{tab:nhr}
    \resizebox{\linewidth}{!}{
    \begin{tabular}{ccccccc}
    \hline
         Method&   MRP~\cite{mrpzhang2017fast} &OSFD~\cite{nhrzhang2020nighttime} &HCD~\cite{hcdwang2024restoring} &FSNet~\cite{FSNet} &FocalNet~\cite{focalnetcui2023focal} &\textbf{ECFNet(Ours)} \\
         \hline
         \hline
        PSNR & 19.93 &21.32 &23.43 &24.35&\underline{25.35} & \textbf{25.51} \\
         SSIM&  0.777 &0.804 &0.953&0.965 &\underline{0.969}&\textbf{0.971} \\
         \hline
    \end{tabular}}
   
\end{table}

\subsubsection{Single-Image Defocus Deblurring}

\begin{figure*}[htb] 
	\centering
	\includegraphics[width=1\linewidth]{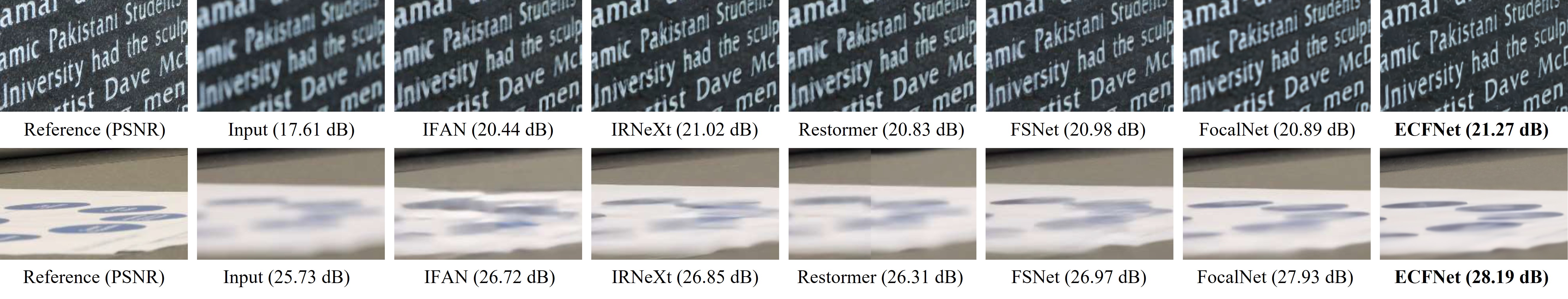}
	\caption{Single-image defocus deblurring results on the DPDD dataset~\cite{DPDNet}.}
	\label{fig:ddpd}
\end{figure*}

We conduct single-image defocus deblurring experiments using the DPDD~\cite{DPDNet} dataset. Table~\ref{tab:deblurd} presents the quantitative comparisons of state-of-the-art defocus deblurring methods. ECFNet outperforms all other methods across all scene categories. Specifically, for the combined scene, ECFNet shows a 0.32 dB improvement over the leading method IRNeXt~\cite{IRNeXt} with up to a 57.9\% reduction in cost, as illustrated in Figure~\ref{fig:param}. Additionally, compared to the Transformer model Restormer~\cite{Zamir2021Restormer}, our ECFNet achieves a significant gain of 0.77 dB on indoor scenes with up to a 78.7\% reduction in cost. The visual results in Figure~\ref{fig:ddpd} demonstrate that our method recovers more details compared to other algorithms.

\begin{table*}[htb]
    \centering
        \caption{Quantitative comparisons with other single-image defocus deblurring methods on the DPDD testset~\cite{DPDNet} (containing 37 indoor and 39 outdoor scenes).}
    \label{tab:deblurd}
    \resizebox{\linewidth}{!}{
    \begin{tabular}{c|cccc|cccc|cccc}
        \hline
    \multicolumn{1}{c|}{} & \multicolumn{4}{c|}{Indoor Scenes}  & \multicolumn{4}{c|}{Outdoor Scenes} & \multicolumn{4}{c}{Combined}
    \\
   Methods & PSNR $\uparrow$ & SSIM $\uparrow$ &MAE $\downarrow$  &LPIPS $\downarrow$ & PSNR $\uparrow$ & SSIM $\uparrow$  &MAE $\downarrow$  &LPIPS $\downarrow$  &  PSNR $\uparrow$ &  SSIM $\uparrow$ &MAE $\downarrow$  &LPIPS $\downarrow$ 
    \\
    \hline
    \hline
    EBDB~\cite{EBDB}& 25.77 &0.772 &0.040 &0.297 &21.25 &0.599 &0.058 &0.373 &23.45 &0.683 &0.049 &0.336
    \\
DMENet~\cite{DMENet}  &25.50 &0.788 &0.038 &0.298 &21.43 &0.644 &0.063 &0.397 &23.41 &0.714 &0.051 &0.349
\\
JNB~\cite{JNB} &26.73 &0.828 &0.031 &0.273 &21.10 &0.608 &0.064 &0.355 &23.84 &0.715 &0.048 &0.315
\\
DPDNet~\cite{DPDNet} & 26.54 &0.816 &0.031 &0.239 &22.25 &0.682 &0.056 &0.313 &24.34 &0.747 &0.044& 0.277
\\
KPAC~\cite{KPAC}& 27.97 &0.852 &0.026 &0.182 &22.62 &0.701 &0.053 &0.269 &25.22 &0.774 &0.040 &0.227
\\
IFAN~\cite{IFAN}& 28.11 &0.861 &0.026 &0.179 &22.76 &0.720 &0.052 &0.254 &25.37 &0.789 &0.039 &0.217
\\
Restormer~\cite{Zamir2021Restormer}& 28.87 &\underline{0.882} &0.025 &\underline{0.145} &23.24 &0.743 &0.050 &\textbf{0.209} &25.98 &0.811 &0.038 &\textbf{0.178}
\\
IRNeXt~\cite{IRNeXt} &\underline{29.22} &0.879 &\underline{0.024} &0.167 &\underline{23.53} &\underline{0.752} &\underline{0.049} &0.244 &\underline{26.30} &\underline{0.814} &\underline{0.037} &0.206
\\
SFNet~\cite{SFNet} &29.16 &0.878 &\underline{0.023} &0.168 &23.45 &0.747 &\underline{0.049} &0.244 &26.23 &0.811 &\underline{0.037} &0.207
\\
FocalNet~\cite{focalnetcui2023focal}& 29.10 &0.876 &\underline{0.024} &0.173 &23.41 &0.743 &\underline{0.049} &0.246 &26.18 &0.808 &\underline{0.037} &0.210
\\
FSNet~\cite{FSNet} &29.14 &0.878 &\underline{0.024} &0.166 &23.45 &0.747 &0.050 &0.246 &26.22 &0.811 &\underline{0.037} &0.207
\\

\hline
\textbf{ECFNet(Ours)}&\textbf{29.64}	&\textbf{0.898}	&\textbf{0.022}	&\textbf{0.142}	&\textbf{23.75}	&\textbf{0.787}	&\textbf{0.047}	&\underline{0.214}	&\textbf{26.62}	&\textbf{0.831}	&\textbf{0.035}	&\underline{0.179}

    \\
    \hline
    \end{tabular}}
\end{table*}

\subsubsection{Image Desnowing}
\begin{figure*}[htb] 
	\centering
	\includegraphics[width=1\linewidth]{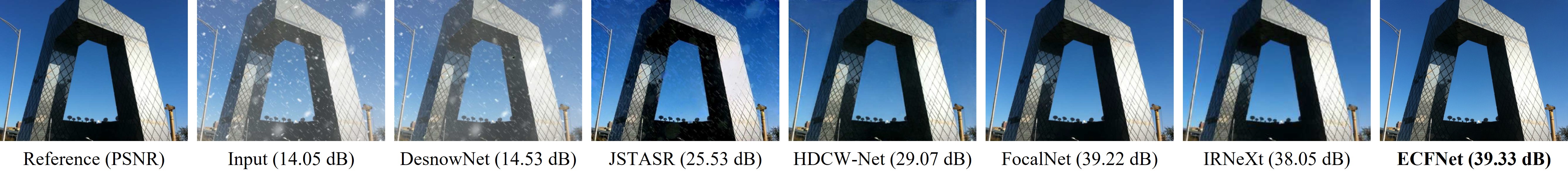}
	\caption{Image desnowing results on the CSD dataset~\cite{HDCW-Netchen2021all}.}
	\label{fig:csd}
\end{figure*}

We compare desnowing performance across three datasets: Snow100K~\cite{desnownet}, SRRS~\cite{JSTASRchen2020jstasr}, and CSD~\cite{HDCW-Netchen2021all}. As shown in Table~\ref{tab:snow}, our ECFNet outperforms other state-of-the-art algorithms. Specifically, ECFNet achieves a 0.70 dB improvement on the SRRS dataset~\cite{JSTASRchen2020jstasr} and a 0.33 dB improvement on the recently proposed CSD dataset~\cite{HDCW-Netchen2021all} compared to IRNeXt~\cite{IRNeXt}. Moreover, compared with MSP-Former~\cite{mspformer10095605}, which is specifically designed for desnowing, our model demonstrates substantial performance gains of 3.87 dB, 1.85 dB, and 0.83 dB on the three datasets, respectively. Figure~\ref{fig:csd} illustrates that ECFNet produces snow-free images without artifacts.

\begin{table}
    \centering
        \caption{Quantitative comparisons with other image  desnowing methods on the three widely used datasets: CSD~\cite{HDCW-Netchen2021all}, SRRS~\cite{JSTASRchen2020jstasr} and Snow100K~\cite{desnownet}.}
    \label{tab:snow}
    \resizebox{\linewidth}{!}{
    \begin{tabular}{c|cc|cc|cc}
    \hline
    \multicolumn{1}{c|}{} & \multicolumn{2}{c|}{CSD}  & \multicolumn{2}{c|}{SRRS} & \multicolumn{2}{c}{Snow100K}
    \\
   Methods & PSNR $\uparrow$ & SSIM $\uparrow$  & PSNR $\uparrow$ & SSIM $\uparrow$ & PSNR $\uparrow$ & SSIM $\uparrow$
   \\
   \hline
   \hline
DesnowNet~\cite{desnownet} &20.13 &0.81 &20.38 &0.84 &30.50 &0.94
\\
CycleGAN~\cite{CycleGANengin2018cycle} &20.98 &0.80 &20.21 &0.74 &26.81 &0.89
\\
All in One~\cite{AllinOneli2020all} &26.31 &0.87 &24.98 &0.88 &26.07 &0.88
\\
JSTASR~\cite{JSTASRchen2020jstasr} &27.96 &0.88 &25.82 &0.89 &23.12 &0.86
\\
HDCW-Net~\cite{HDCW-Netchen2021all} &29.06 &0.91 &27.78 &0.92 &31.54 &\underline{0.95}
\\
TransWeather~\cite{TransWeathervalanarasu2021transweather} &31.76 &0.93 &28.29 &0.92 &31.82 &0.93
\\
 NAFNet~\cite{chen2022simple} &33.13 &0.96 &29.72 &0.94 &32.41 &\underline{0.95}
\\
MSP-Former~\cite{mspformer10095605} & 33.75 &0.96 &30.76 &0.95 &33.43& \textbf{0.96}
\\
FocalNet~\cite{focalnetcui2023focal} &37.18 &\textbf{0.99} &31.34 &\textbf{0.98} &33.53 &\underline{0.95}
\\
IRNeXt~\cite{IRNeXt} &37.29 &\textbf{0.99} &\underline{31.91} &\textbf{0.98} &\underline{33.61} &\underline{0.95}
\\
         \hline
\textbf{ECFNet(Ours)} &\textbf{37.62} &\underline{0.98}  &\textbf{32.61} &\underline{0.96} &\textbf{34.26} & \textbf{0.96}
         \\
         \hline
    \end{tabular}}
\end{table}

\subsection{Ablation Studies}
We conduct ablation studies on the Snow100K dataset~\cite{desnownet} to demonstrate the effectiveness of our modules. We use the three-scale NAFNet~\cite{chen2022simple} as the baseline network and perform a step-by-step ablation study by successively integrating the proposed modules into the baseline.

\begin{table}
    \centering
      \caption{Ablation studies for different components of ECFNet on
the Snow100K dataset~\cite{desnownet}.}
    \label{tab:abl}
    \begin{tabular}{ccccc}
    \hline
         &MSBlock& MSSFBlock & SFAM  & PSNR
         \\
         \hline
         (a)& \ding{56}&  \ding{56}&   \ding{56}& 32.37
         \\
         (b)& \ding{52}&  \ding{56}&    \ding{56}& 32.51
         \\
         (c)& \ding{56}&  \ding{52}&   \ding{56}&  32.60
         \\
         (d)& \ding{56}&  \ding{56}&  \ding{52}&   33.82
         \\
         (e)& \ding{52}&  \ding{52}&  \ding{56}&   33.04
         \\
         (f)& \ding{52}&  \ding{56}&  \ding{52}&  33.99
         \\
         (g)& \ding{56}&  \ding{52}&  \ding{52}&  34.11
         \\
         (h)& \ding{52}&  \ding{52}&  \ding{52}& \textbf{34.26}
         \\
         \hline
    \end{tabular}
\end{table}

\textbf{Effects of Individual Components.} 
As shown in Table~\ref{tab:abl}(a), the baseline achieves a PSNR of 32.41 dB. When the original block at the first scale of the encoder-decoder is replaced with MSBlock, as seen in Table~\ref{tab:abl}(b), the model consistently achieves better results with an improvement of 0.14 dB in PSNR. Next, we replace all the original blocks with MSSFBlock, resulting in a performance improvement of 0.23 dB PSNR, as shown in Table~\ref{tab:abl}(c). When we substitute the original bottleneck with SFAM (Table~\ref{tab:abl}(d)), the performance increases by 1.45 dB PSNR. 
Regardless of how we combine these three modules, each combination results in a corresponding performance improvement. When all modules are used together (Table~\ref{tab:abl}(h)), our model achieves a 1.89 dB boost over the original baseline (Table~\ref{tab:abl}(a)).

\begin{table}
    \centering
      \caption{Design choices for SFAM. SFAM\_b$i$ denotes the $i$ th branch.}
    \label{tab:ablsfam}
    \begin{tabular}{cccccc}
    \hline
         Method& SFAM\_b1 & SFAM\_b2 & SDAM & FDAM & PSNR
         \\
         \hline
         (a)& \ding{56}&  \ding{56}&   \ding{56} &   \ding{56}& 33.04
         \\
         (b)& \ding{56}&  \ding{56}&   \ding{52} &   \ding{56}& 33.82
         \\
         (c)& \ding{56}&  \ding{56}&   \ding{56} &   \ding{52}& 33.31
         \\
         (d)& \ding{52}&  \ding{56}&  \ding{52} &    \ding{52}& 34.15
         \\
         (e)& \ding{56}&  \ding{52}&  \ding{52} &    \ding{52}& 34.02
         \\
         (f)& \ding{52}&  \ding{52}&  \ding{52}&  \ding{52}& \textbf{34.26}
         \\
         \hline
    \end{tabular}
\end{table}

\begin{table}
    \centering
      \caption{Comparisons with other attention mechanisms.}
    \label{tab:ablatten}
    \begin{tabular}{ccc}
    \hline
         &Method& PSNR
         \\
         \hline
          (a)&SAM~\cite{Zamir2021MPRNet} &32.22
         \\
          (b)&SSM~\cite{focalnetcui2023focal} & 33.45
          \\
          (c)&SDAM (Ours) & \textbf{33.82}
          \\
        \hline
          (d)&FSM~\cite{focalnetcui2023focal} & 33.10
          \\
          (e)&FDAM (Ours) &\textbf{33.31}
          \\
          \hline
          (f)&DSM~\cite{focalnetcui2023focal}&33.64
          \\
          (g)&SDAM+SDAM & 34.02
          \\
          (h)&FDAM+FDAM & 33.49
          \\
          (i)&SDAM+FDAM (Ours)&\textbf{34.15}
         \\
         \hline
    \end{tabular}
\end{table}

\begin{figure*}[htb] 
	\centering
	\includegraphics[width=1\linewidth]{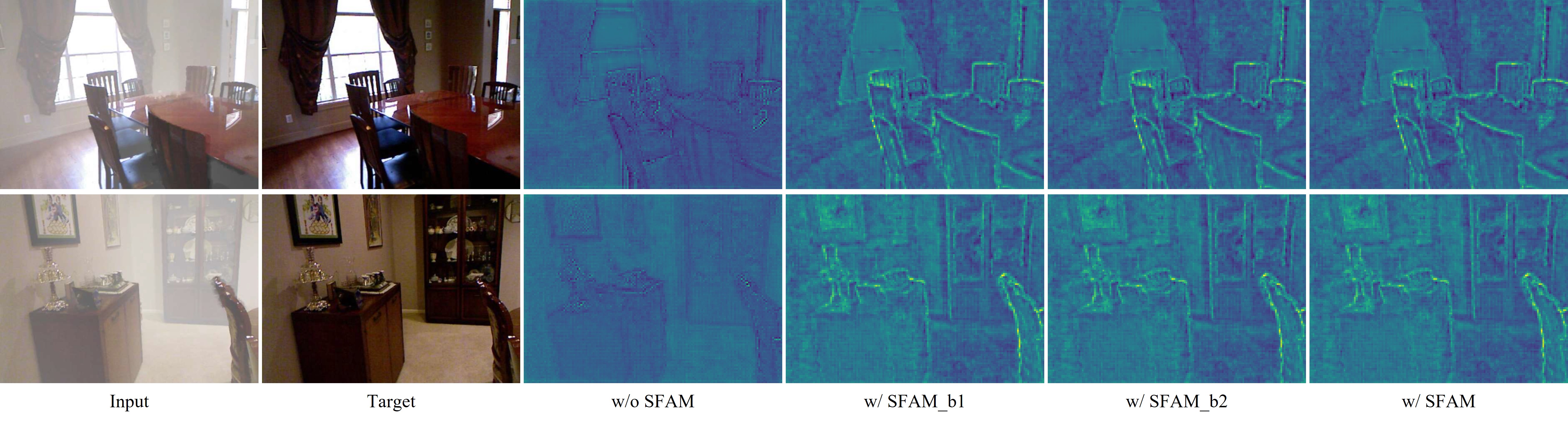}
	\caption{Visual results of SFAM.}
	\label{fig:sfamabl}
\end{figure*}

\begin{figure*}[htb] 
	\centering
	\includegraphics[width=1\linewidth]{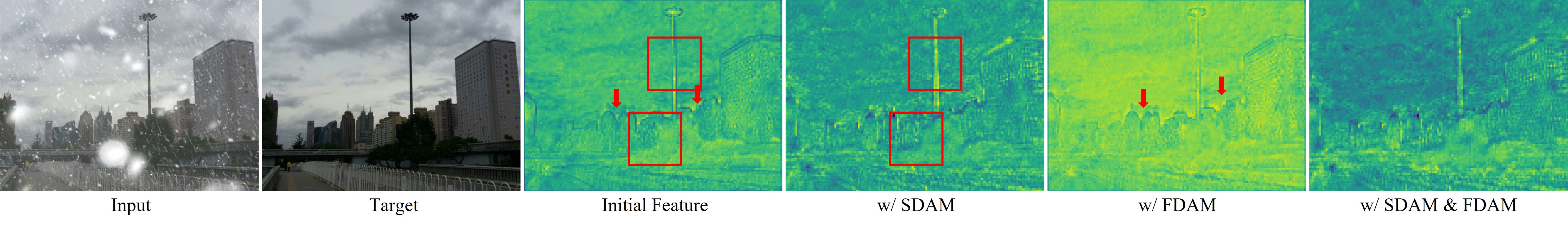}
	\caption{Effects of our SFAM.}
	\label{fig:sfamsi}
\end{figure*}

\begin{figure*}[htb] 
	\centering
	\includegraphics[width=1\linewidth]{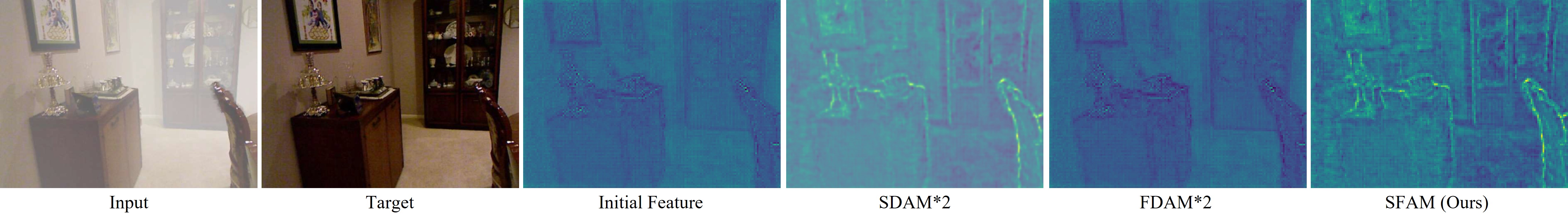}
	\caption{Comparisons between different designs of SFAM.}
	\label{fig:sfamco}
\end{figure*}

\textbf{Design Choices for SFAM.} 
To assess the capability of SFAM, we conducted several experimental designs. We use the model in Table~\ref{tab:abl}(e) as the base model and add the corresponding modules incrementally. As shown in Table~\ref{tab:ablsfam}, adding SDAM (Table~\ref{tab:ablsfam}(b)) and FDAM (Table~\ref{tab:ablsfam}(c)) results in accuracy gains of 0.76 dB and 0.27 dB over the base model (Table~\ref{tab:ablsfam}(a)), respectively. When SDAM and FDAM are combined, the performance improvement is even more significant (Table~\ref{tab:ablsfam}(d)(e)). We also found that the order of combining SDAM and FDAM impacts the results. For instance, in SFAM\_b1, SDAM is used first to locate degenerate areas, followed by FDAM to enhance high-frequency signals, while SFAM\_b2 uses the opposite order. When both branches are used simultaneously and their results are fused (Table~\ref{tab:ablsfam}(f)), the performance improves further, showing a 1.22 dB gain over the base model (Table~\ref{tab:ablsfam}(a)).

Furthermore, we demonstrate the superiority of our approach by replacing our proposed SDAM and FDAM with other attention mechanisms, as shown in Table~\ref{tab:ablatten}. Replacing our SDAM with SAM~\cite{Zamir2021MPRNet} and SSM~\cite{focalnetcui2023focal} results in performance drops of 1.60 dB and 0.37 dB, respectively. Similarly, replacing our FDAM with FSM~\cite{focalnetcui2023focal} decreases performance by 0.21 dB. These results highlight the effectiveness of our design. Moreover, when we deploy two identical modules (Table~\ref{tab:ablatten}(g)(h)), the models achieve better performance than when using a single one (Table~\ref{tab:ablatten}(c)(e)). Our approach, employing both spatial and frequency selection mechanisms, yields the best performance (Table~\ref{tab:ablatten}(i)). It also improves performance by 0.51 dB compared to the existing attention mechanism DSM~\cite{focalnetcui2023focal}.

Finally, to demonstrate the superiority of our method, we provide visualization comparisons between different designs. As depicted in Figure~\ref{fig:sfamabl}, our SFAM emphasizes hard-to-recover regions, resulting in more fine details compared to the model without SFAM. In Figure~\ref{fig:sfamsi}, SDAM helps the model suppress simple regions and focus more on severe degradation areas, such as the building in the bottom red area. FDAM further enhances edge signals by removing low-frequency information, as indicated by the red arrows. In Figure~\ref{fig:sfamco}, using two SDAMs, the model focuses more on degradation regions, while the version with two FDAMs pays more attention to edge signals. 

\begin{table}
    \centering
      \caption{The number of branches in MSBlock. The number indicates the downsampling rate of a branch.}
    \label{tab:ablmsb}
    \begin{tabular}{ccccccc}
    \hline
         Method& 1&2&4 &8 & PSNR
         \\
         \hline
         (a)& \ding{52}&\ding{56}&  \ding{56}&   \ding{56} & 34.11
         \\
         (b)& \ding{52}&\ding{52}&  \ding{56}&   \ding{56} & 34.20
         \\
         (c)& \ding{52}&\ding{52}&  \ding{52}&   \ding{56}& \textbf{34.26}
         \\
         (d)& \ding{52}&\ding{52}&  \ding{52}&  \ding{52} & \textbf{34.26}

         \\
         \hline
    \end{tabular}
\end{table}

\begin{table}
    \centering
      \caption{The effect of SCABlock and  MSFBlock.}
    \label{tab:ablsca}
    \begin{tabular}{cccc}
    \hline
         Method& SCABlock& MSFBlock& PSNR
         \\
         \hline
         (a)& \ding{56}&  \ding{56}&   33.82
         \\
         (b)& \ding{52}&  \ding{56} & 33.99
         \\
         (c) & \ding{56}&\ding{52}&34.07
         \\
         (d)& \ding{52}&\ding{52}&\textbf{34.26}

         \\
         \hline
    \end{tabular}
\end{table}

\textbf{Design Choices for MSBlock.} 
We investigate the impact of the number of branches in MSBlock. As demonstrated in Table~\ref{tab:ablmsb}, employing more branches results in better performance. Specifically, when using a single branch with a downsampling rate of 2, the model achieves a gain of 0.09 dB over the baseline. With three branches, the model shows a 0.15 dB improvement. However, adding more branches does not further enhance performance. This is because with three branches, our model already effectively learns multi-scale representations.

\textbf{Effect of SCABlock and  MSFBlock.} 
We investigate the impact of incorporating SCABlock and MSFBlock into our model. As shown in Table~\ref{tab:ablsca}, SCABlock and MSFBlock independently enhance performance by 0.17 dB and 0.25 dB, respectively, compared to the baseline. When both blocks are utilized together, they result in an even greater performance boost of 0.44 dB. This significant improvement demonstrates the effectiveness of our proposed design.

\section{Conclusion}
Different regions in a corrupted image experience varying degrees of degradation. Addressing this, we propose an efficient and effective framework for image restoration, called ECFNet. ECFNet consists of two key components: the spatial and frequency attention mechanism (SFAM) and the multi-scale block (MSBlock). The SFAM utilizes the spatial domain attention module (SDAM) to identify degraded regions and the frequency domain attention module (FDAM) to enhance high-frequency signals, thereby highlighting crucial features for restoration. To support multi-scale representation learning and capture global dependencies, we design the MSBlock with three scale branches, each containing multiple simplified channel attention blocks (SCABlocks) and a multi-scale feedforward block (MSFBlock). Extensive experimental results demonstrate that ECFNet outperforms state-of-the-art methods.

\bibliographystyle{IEEEtran}
\bibliography{refbib}
\vfill
\end{document}